\documentclass[svproc]{llncs}

\usepackage{physics}
\usepackage{enumitem}
\usepackage{eso-pic}
\usepackage{physics}
\usepackage{xcolor}
\usepackage[nolist,nohyperlinks]{acronym}
\usepackage{glossaries}
\usepackage{cite}
\usepackage{orcidlink}
\usepackage{tabularx}
\usepackage{pifont}
\usepackage{pstricks}
\usepackage{verbatim}
\usepackage{psfrag}
\usepackage[english]{babel}
\usepackage{amsmath}
\usepackage{amssymb}
\usepackage[font=small]{caption}
\usepackage{fancybox}
\usepackage{steinmetz}
\usepackage{subcaption}
\usepackage{bodegraph}
\usepackage{gensymb}
\usepackage{algorithm,algorithmic}
\usepackage{amsfonts}
\usepackage{dsfont}
\usepackage{siunitx}
\usepackage{lscape}
\usepackage{enumerate}
\usepackage{lmodern}
\usepackage[T1]{fontenc}
\usepackage[framed]{matlab-prettifier}
\usepackage{booktabs}
\usepackage{mathtools}
\usepackage{color} 
\usepackage{xcolor,colortbl}
\usepackage{colortbl}
\usepackage{multirow, hhline}
\usepackage{url}
\usepackage{bm}
\usepackage{multicol}
\usepackage{supertabular}
\usepackage{graphicx}
\usepackage{hyperref}
\usepackage[utf8]{inputenc}
\usepackage{tikz}
\usepackage{tabularx}
\usepackage{array}
\usepackage{threeparttable}
\usepackage{multirow}
\usepackage{placeins}

\newacronym{fsm}{FSM}{Finite State Machine}
\newacronym{slerp}{SLERP}{Spherical Linear Interpolation}
\newacronym{vlms}{VLMs}{Vision-Language Models}
\newacronym{llms}{LLMs}{Large Language Models}
\newacronym{sus}{SUS}{System Usability Scale}
\newacronym{pHRI}{pHRI}{physical Human-Robot Interaction}
\definecolor{Gray}{gray}{0.9}

\definecolor{verdedd}{rgb}{0.0, 0.5, 0.0}

\definecolor{myorangenew}{rgb}{1 0.49 0}
\definecolor{mygreennew}{rgb}{0.31 0.78 0.47}
\definecolor{mygreynew}{rgb}{0.6 0.6 0.6}
\definecolor{bluD}{rgb}{0.09, 0.45, 0.81}
\definecolor{bluD_l}{rgb}{0.29, 0.65, 1}
\definecolor{nero}{rgb}{0, 0, 0}
\definecolor{bianco}{rgb}{1, 1, 1}

\definecolor{bluG}{cmyk}{100 85 0 0}
\definecolor{verdeG}{cmyk}{75 0 100 0}

   \def\p{{\boldsymbol p}}

 \def\x{{\boldsymbol x}}

 \def\0{{\boldsymbol 0}}

\DeclareMathSymbol{\boldmu}{\mathord}{letters}{15}
\newcommand{\T}[2]{{}^{\scriptscriptstyle #1}\mathbf{T}_{\scriptscriptstyle #2}}
\newcommand{\R}[2]{{}^{\scriptscriptstyle #1}\mathbf{R}_{\scriptscriptstyle #2}}
\newcommand{\pp}[2]{{}^{\scriptscriptstyle #1}\boldsymbol{p}_{\scriptscriptstyle #2}}

\begin{document}

\AddToShipoutPictureBG*{%
  \AtPageUpperLeft{%
    \put(0,-45){%
      \makebox[\paperwidth][c]{%
        \parbox{0.90\paperwidth}{%
          \centering
          \scriptsize
          \textbf{Preprint version.}
          This paper has been accepted for publication in the
          \textit{Springer Proceedings in Advanced Robotics} as part of the
          proceedings of the 19th International Workshop on
          Human-Friendly Robotics (HFR 2026), Trento, Italy, July 16--17, 2026.
          The final authenticated version will be available online at
          Springer Nature Link.
        }%
      }%
    }%
  }%
}

\title{Receiver-Centered Robot-to-Human Handover with Grasp-Aware Object Orientation}

\titlerunning{Receiver-Centered Handover in Human-Robot Interaction}

\author{Federico Biagi \and
Dario Onfiani \and Simone Silenzi \and Luigi Biagiotti}

\authorrunning{Biagi et al.}

\institute{Department of Engineering "Enzo Ferrari", University of Modena and Reggio Emilia, Italy.
\email{\{federico.biagi, dario.onfiani, simone.silenzi, luigi.biagiotti\}@unimore.it}}

\maketitle

\begin{abstract}
Collaborative robots are increasingly sharing workspaces with human operators, making tool handover a frequent and safety-critical micro-interaction. However, traditional static handovers often lead to awkward grasps when handling asymmetric industrial tools. This paper presents a receiver-centered voice-driven adaptive handover system for mechanical tools, built on a Franka cobot. Using an LLM for intention recognition and MediaPipe for real-time 3D hand tracking, the framework dynamically adjusts the end-effector's orientation to present tools in an ergonomically optimal, handle-first pose. A within-subjects study compared this adaptive approach with an object-agnostic static baseline. The results demonstrate that the adaptive system reduces the grasp delay for asymmetric tools, improving the fluency of the interaction. Furthermore, the adaptive strategy improved specific trust-related perceptions, particularly motion predictability and perceived task simplicity.

\end{abstract}

\section{Introduction}
\label{sec:Introduction}
In contemporary manufacturing, the shared workspace between human operators and cobots makes the handover of mechanical tools a highly common and safety-critical procedure. Fluent exchange reduces cycle time and prevents the need for repeated awkward reaches; conversely, a poorly executed handover disrupts cognitive flow and may pose a risk of injury, particularly when handling tools such as hammers or screwdrivers. Owing to the functional asymmetry of industrial tools (e.g., handle versus head, shaft versus sharp end), effective handover requires the robot to present the correct functional part of the object, with an appropriate orientation, at the right location and at the right moment.\newline
%
%
Object handover is commonly framed as a joint action comprising pre-handover communication, approach, and a physical exchange regulated by coupled grip and load forces \cite{strabala2013towards,ortenzi2021object}. Adaptive givers synchronize spatial and temporal decisions with the receiver, selecting configurations that human partners reliably accept \cite{cakmak2011human}. Recent contributions moved from hand-designed controllers to continuous adaptation via contextual policy search \cite{kupcsik2017learning} and reactive grasp-prediction pipelines \cite{yang2021reactive,wang2024contacthandover}. For the Franka Panda, hand-tracked adaptation via MediaPipe \cite{mediapipe_2020} is consistently preferred over static baselines \cite{kappler2023optimizing}.\newline
%
%
An active research area has focused on the analysis and synthesis of approach trajectories to the handover position.
Two classical families dominate handover motion design. Minimum-jerk profiles model natural human reaching with bell-shaped velocity curves \cite{flash1985coordination,pan2019fast}, a pattern confirmed by human-human approach-and-handover studies in which transfer occurs near the midpoint between partners \cite{basili2009investigating}. Demonstration-based Dynamic Movement Primitives \cite{ijspeert2013dynamical,BRAGLIA2025} offer a second family, while strictly geometric Bézier paths \cite{simba2017vision} allow trivial online re-parameterization and represent our choice for approach trajectory. Because legibility and predictability can conflict \cite{dragan2013legibility}, we deliberately prioritize predictable motion: orientation alignment is engaged only once the user's hand enters the handover area, keeping the robot's spatial target and functional intent unambiguous.
\subsection{Evaluation framework and positioning of our work}
\label{subsec:EvaluationFrameworkAndPositioning}
Mixed-methods evaluation combining objective signals with validated self-reports is standard in handover research \cite{hoffman2019evaluating}. Handover time serves as a dominant behavioral proxy for user hesitancy and ergonomic comfort. For cognitive load, physiological signals are widely utilized; specifically, spontaneous blink rate is an unobtrusive, validated attentional index \cite{stern1984endogenous,chen2014using} recently confirmed in cobot-assisted assembly \cite{pluchino2023advanced}, though its real-time use to discriminate handover strategies remains unexplored. To capture the user's subjective experience, we employ the NASA Task Load Index (NASA-TLX) \cite{hart2006nasa} and the Human-Robot Trust Scale \cite{Charalambous_2015}.\newline
Against this landscape, our framework advances the state of the art along three axes. First, we extend adaptation \cite{kupcsik2017learning,yang2021reactive} to mechanical tools via a unified handle-first strategy. Second, we integrate a hands-free Large-Language-Model (LLM) voice interface. To guarantee deterministic safety in collaborative cells, the LLM is strictly bounded to a narrow classification role \cite{brohan2023can,ao2024behavior}, selecting an object label from a fixed set, while motion planning remains controller-based. Third, we uniquely triangulate the aforementioned objective and subjective metrics to comprehensively assess interaction fluency, attentional engagement, and relational trust. A video demonstration of the proposed framework can be viewed at \url{https://youtube.com/shorts/EC_4DCmi4XA}
\section{Study Description}
\label{sec.Study_Description}
%
%
\subsection{Participants}
\label{subsec:participants}
A total of 15 right-handed volunteers (12 male, 3 female; majority under 35 years of age) took part in the experiment. The group was evenly divided regarding prior exposure to robotics, with approximately half of the subjects reporting previous experience. All experimental procedures were approved by the local institutional ethics board and conducted in strict adherence to the principles of the Declaration of Helsinki. Written informed consent was obtained from every individual prior to their involvement in the study.
\subsection{Robotic setup}
\label{subsec.Framework}
\begin{figure}[t]
    \centering
    \includegraphics[width=0.7\columnwidth]{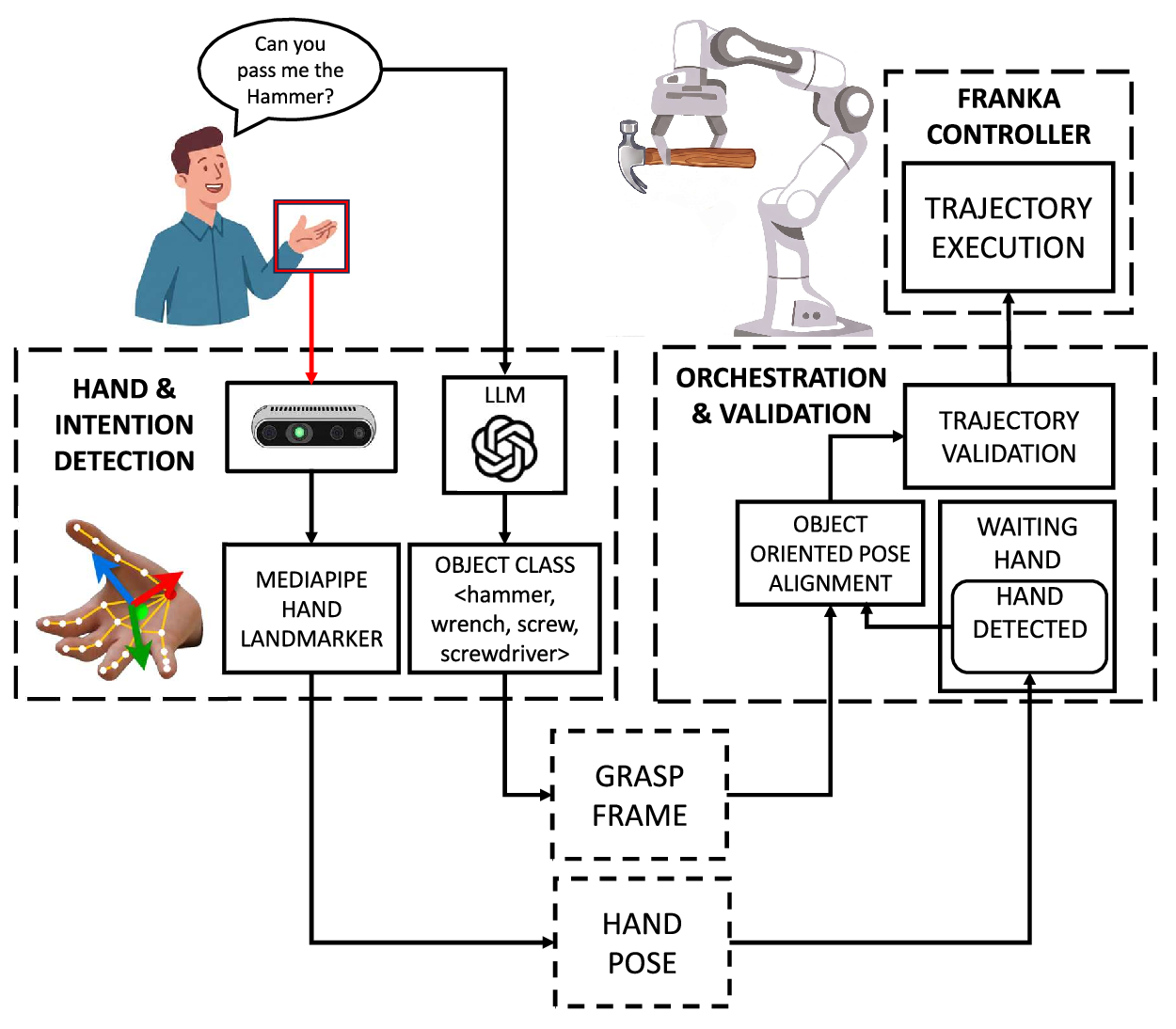}
    \caption{Schematic of the proposed Adaptive handover framework. The system is divided into three blocks: \textit{Hand \& Intention Detection}: extracts 3D hand position and orientation using MediaPipe, the user request is classified through an LLM into one of the available objects; \textit{Orchestration and Validation} implements the handover Finite State Machine (FSM) and computes the object-oriented pose alignment taking both the hand pose and the target grasp frame as input, it then validates the trajectory; \textit{Franka Controller} executes the safe and smooth trajectory}
    \label{fig:framework}
\end{figure}

As depicted in Fig. \ref{fig:framework}, the handover framework  consists of a Franka Emika Panda cobot and an Intel RealSense D435i camera. The hand detection block merges 2D keypoints and depth data via MediaPipe Hands \cite{mediapipe_2020} for 3D hand pose detection. An LLM classifies the user request to make the cobot gather the correct object. The extracted data enables a dynamical alignment of the end-effector to the user's hand. Finally, the target trajectory is validated through an inverse kinematics solver to prevent singularities and joint limit violations.

\subsection{Kinematics and Hand Target Computation}
The employed architecture fuses distinct data sources to robustly estimate the human receiver's pose. The system defines four primary coordinate frames: the Camera frame ($\mathcal{F}_{C}$), the Grasp frame ($\mathcal{F}_{G}$), the Robot Base frame ($\mathcal{F}_{B}$) and the Hand frame ($\mathcal{F}_{H}$). To represent poses and compute spatial relationships concisely, we adopt homogeneous transformation matrices $\T{}{} \in SE(3)$.
To extract the hand pose in the camera frame, $\T{C}{H}$, we propose a hybrid analytical approach that mitigates the intrinsic noise of depth sensors. 
First, the hand's spatial position $\pp{C}{H} \in \mathbb{R}^{3}$ is obtained by back-projecting the 2D MediaPipe landmarks using the camera's aligned depth map and intrinsic parameters. Specifically, let $\pp{C}{W}$ and $\pp{C}{M}$ be the 3D coordinates of the wrist and the middle finger metacarpophalangeal (MCP) joint, respectively. An Exponential Moving Average (EMA) filter is applied to these 3D points to eliminate spatial jitter.
The center of the hand frame is defined as a point along the line connecting the wrist and the base of the middle finger, i.e.,
\[
    \pp{C}{H} = \pp{C}{W} + \lambda (\pp{C}{M} - \pp{C}{W})
\]
where $\lambda \in [0, 1]$ is a tunable parameter controlling the palm ratio.

\begin{figure}[tbp]
    \vspace{-30mm}
    \centering
    \begin{tikzpicture}
        \node[anchor=south west, inner sep=0] (img) at (0,0)
            {\includegraphics[width=0.85\linewidth]{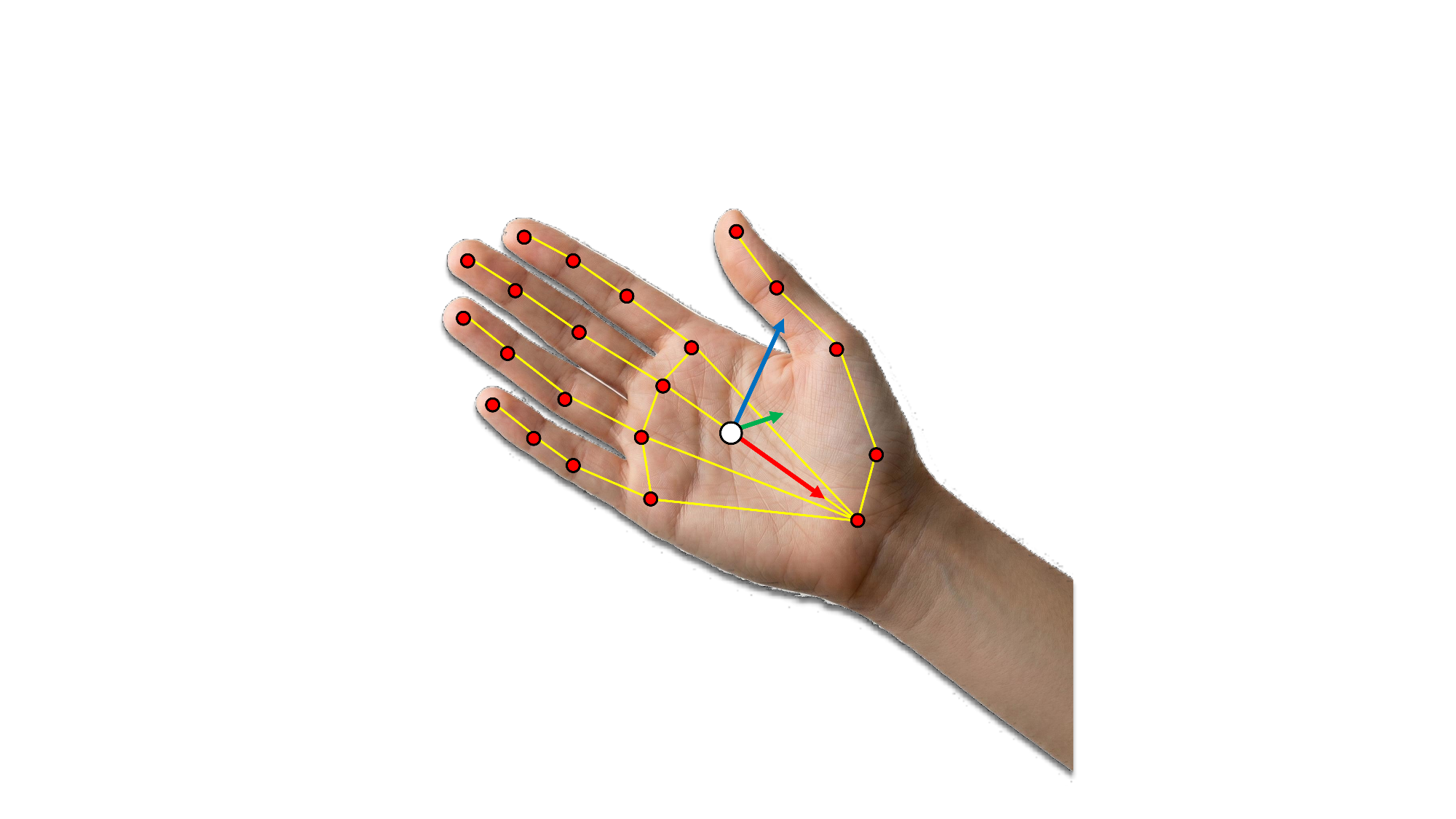}};

        \begin{scope}[x={(img.south east)}, y={(img.north west)}]
            \node[font=\fontsize{8pt}{9pt}\selectfont] at (0.600,0.330) {$\boldsymbol{w}$};
            \node[font=\fontsize{8pt}{9pt}\selectfont] at (0.620,0.435) {$\boldsymbol{t}$};
            \node[font=\fontsize{8pt}{9pt}\selectfont] at (0.475,0.615) {$\boldsymbol{i}$};
            \node[font=\fontsize{8pt}{9pt}\selectfont] at (0.450,0.565) {$\boldsymbol{m}$};
            \node[font=\fontsize{8pt}{9pt}\selectfont] at (0.440,0.495) {$\boldsymbol{r}$};
            \node[font=\fontsize{8pt}{9pt}\selectfont] at (0.445,0.420) {$\boldsymbol{p}$};

            \node[text=red, font=\fontsize{9pt}{10pt}\selectfont] at (0.550,0.380) {$\boldsymbol{x_h}$};
            \node[text=green!50!black, font=\fontsize{9pt}{10pt}\selectfont] at (0.550,0.5) {$\boldsymbol{y_h}$};
            \node[text=blue, font=\fontsize{9pt}{10pt}\selectfont] at (0.550,0.6) {$\boldsymbol{z_h}$};
        \end{scope}
    \end{tikzpicture}
    \caption{Hand landmarks and hand reference frame.}
    \label{fig:hand_landmarks_axes}
\end{figure}
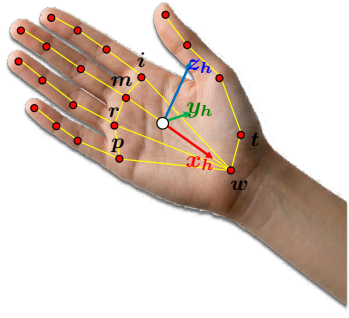
\begin{figure}[t]
    \centering
    \includegraphics[width=\linewidth]{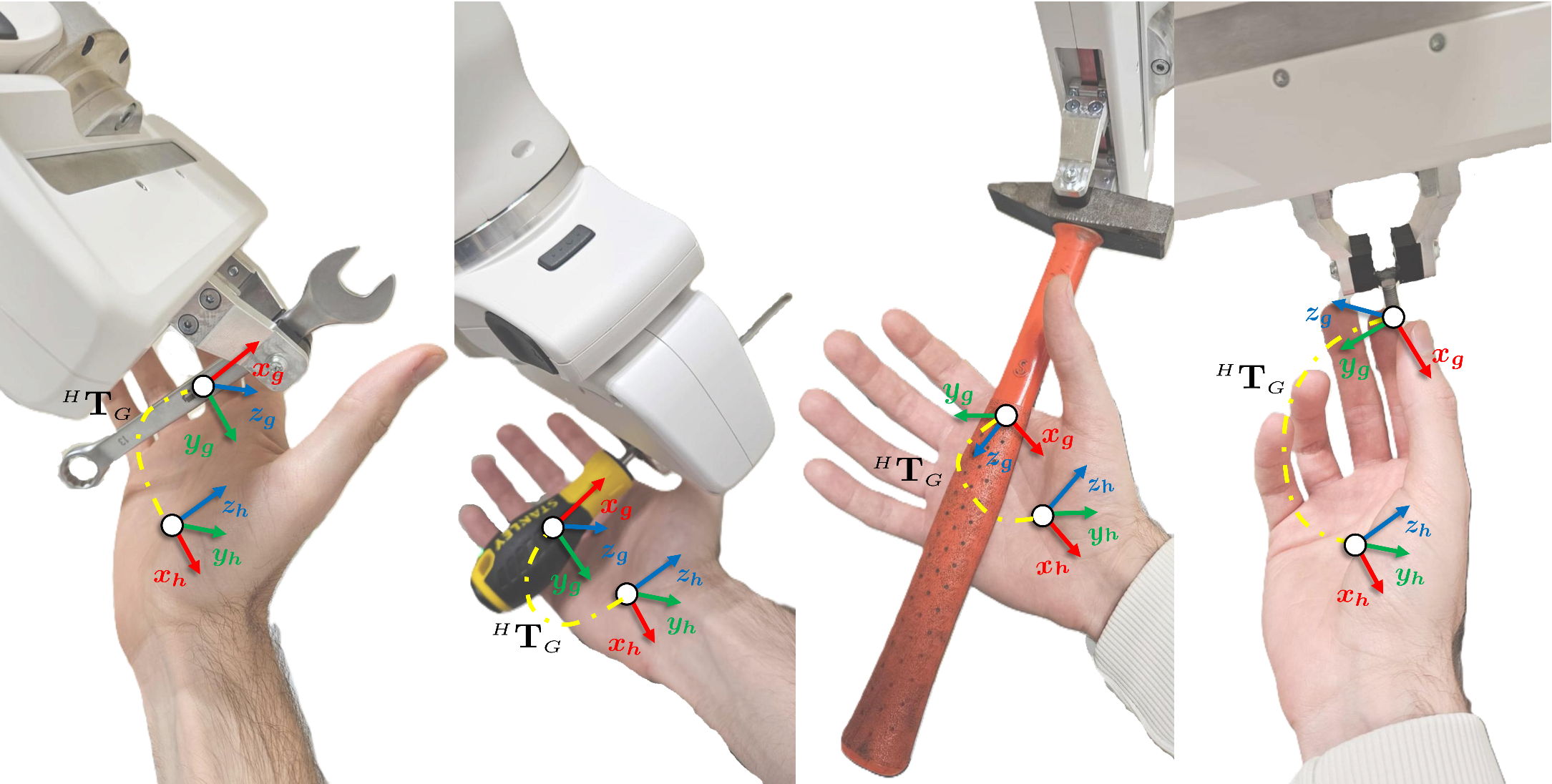}
    \caption{Representation of the grasping poses for different handover tasks. Each image illustrates the effect of the constant homogeneous transformation $\T{H}{G}$ specific to the scenario.}
    \label{fig:grasping_pose}
\end{figure}

To compute the hand's orientation $\R{C}{H}$ without relying on the noisy 3D point cloud, we leverage MediaPipe Hand Landmarks, which provide a scale-normalized kinematic model of the hand; see Fig. \ref{fig:hand_landmarks_axes}. Let $\boldsymbol{w}, \boldsymbol{i}, \boldsymbol{m}, \boldsymbol{r}, \boldsymbol{p} \in \mathbb{R}^{3}$ denote the local 3D coordinates of the wrist, index MCP, middle MCP, ring MCP, and pinky MCP, respectively. We construct a local orthonormal basis corresponding to the hand's attitude. 
The dorsal axis ($\hat{\boldsymbol{y}}$) is defined as the unit normal to the plane spanned by the wrist, index, and pinky joints:
\begin{equation}
    \hat{\boldsymbol{y}} = \frac{(\boldsymbol{p} - \boldsymbol{w}) \times (\boldsymbol{i} - \boldsymbol{w})}{\norm{(\boldsymbol{p} - \boldsymbol{w}) \times (\boldsymbol{i} - \boldsymbol{w})}}
\end{equation}
The longitudinal axis ($\hat{\boldsymbol{x}}$), pointing towards the wrist, is computed orthogonally to both $\hat{\boldsymbol{y}}$ and the transversal vector connecting the ring and middle MCPs:
\begin{equation}
    \hat{\boldsymbol{x}} = \frac{\hat{\boldsymbol{y}} \times (\boldsymbol{m} - \boldsymbol{r})}{\norm{\hat{\boldsymbol{y}} \times (\boldsymbol{m} - \boldsymbol{r})}}
\end{equation}
Finally, the lateral axis ($\hat{\boldsymbol{z}}$) completes the right-handed reference frame:
\begin{equation}
    \hat{\boldsymbol{z}} = \frac{\hat{\boldsymbol{x}} \times \hat{\boldsymbol{y}}}{\norm{\hat{\boldsymbol{x}} \times \hat{\boldsymbol{y}}}}
\end{equation}
The resulting rotation matrix $\R{C}{H} = \begin{bmatrix} \hat{\boldsymbol{x}} & \hat{\boldsymbol{y}} & \hat{\boldsymbol{z}} \end{bmatrix}$ and the position vector $\pp{C}{H}$ are embedded into a single homogeneous transformation matrix representing the hand pose in the camera frame:
\begin{equation}
    \T{C}{H} = 
    \begin{bmatrix}
        \R{C}{H} & \pp{C}{H} \\
        \0_{1 \times 3} & 1
    \end{bmatrix}
\end{equation}
Utilizing a static hand-eye calibration matrix $\T{B}{C}$, which describes the transformation from the camera to the robot base, the hand's spatial pose is transformed into the robot's base frame through a direct matrix multiplication:
\begin{equation}
    \T{B}{H} = \T{B}{C}\T{C}{H}
\end{equation}
For each object, the grasp reference frame (coinciding with the object frame during handover) is defined with respect to the hand frame to ensure task-dependent ergonomic optimization. This is achieved by combining a displacement along the palm axis with a task-specific orientation.\\
These effects are encoded in a constant homogeneous transformation $\T{H}{G}$ (specific to each scenario), as illustrated in Fig.~\ref{fig:grasping_pose}, which captures both the ergonomic offset and the desired grasp orientation to ensure a seamless grasp. The parameters of $\T{H}{G}$ are constant for each tool class and were empirically derived prior to the main experiment during an informal pilot phase. In this phase, a small group of users briefly held each tool to demonstrate their natural receiving grip.
The optimal grasping pose in the robot base frame, $\T{B}{G}$, is then computed as:
\begin{equation}
    \T{B}{G} = \T{B}{H}\T{H}{G}
    \nonumber
\end{equation}

Finally, the target pose for the object required to execute the handover is obtained as:
\begin{equation}
    \hat{\T{B}{O}} = \T{B}{G}.
    \label{eq:TargetPose}
\end{equation}
During the physical execution of the task, the robot's control system aims to align the actual object pose, denoted as $\T{B}{O}$, with this estimated target $\hat{\T{B}{O}}$. The actual pose $\T{B}{O}$ is continuously computed by combining the forward kinematics of the robot arm with the constant transformation that defines the relative pose between the gripper and the grasped object.
\subsection{Robot Control and Trajectory Generation}
\label{subsec.control_traj}
\begin{figure}[tb]
    \centering
    \includegraphics[width=0.65\columnwidth]{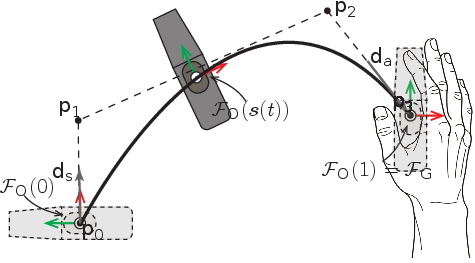}
    \caption{Schematic representation of the cubic Bézier trajectory generated for the handover task of an hammer. The geometric path is shaped by four control points ($\p_0$,$\p_1$,$\p_2$,$\p_3$), imposing the start ($\boldsymbol{d}_{s}$) and approach ($\boldsymbol{d}_{a}$) directions to ensure a smooth, natural motion toward the final grasp frame ($\mathcal{F}_{G}$).}
    \label{fig:bezier}
\end{figure}

The robot's control architecture is developed in ROS2. The main script evaluates a \acrfull{fsm} at a fixed loop rate to autonomously manage both the handover and pick up of objects. The system operates in two main modalities:
\begin{itemize}
\item \textit{Baseline Handover:} The robot tracks the user's hand position but employs a constant, task-agnostic relative transformation $\T{H}{G}$ for all objects. As a result, while the spatial position adapts to the user's hand, the object is always presented with the same default orientation relative to the palm, regardless of its shape or intended use.
\item \textit{Adaptive Handover:} The robot reaches the user's hand with the optimal target pose $\hat{\T{B}{O}}$ defined in \eqref{eq:TargetPose}. Here, the transformation $\T{H}{G}$ is actively modulated based on the specific downstream task, ensuring the object is presented with an ergonomically optimal orientation for seamless use.
\end{itemize}
While the user's hand pose is monitored continuously via the RGB-D stream, the controller commits to a target only when the detected hand remains spatially stable (translation < 5 cm, rotation < 0.26 rad) within the handover workspace for at least 2 s. At this instant, a single target pose is published and held fixed for the entire approach. 
The object pose trajectory is derived by decoupling the geometric path (for both position and orientation) from the motion law:
\begin{equation}
    \T{B}{O}(s) =
    \begin{bmatrix}
        \R{B}{O}(s) & \pp{B}{O}(s) \\
        \0_{1 \times 3} & 1
    \end{bmatrix}
    \nonumber
\end{equation}
with $s(t) \in [0,1]$.\\
To generate a smooth and ergonomically predictable handover motion, the position $\pp{B}{O}(s)$ is defined as a cubic B\'ezier curve in Cartesian space \cite{biagiotti2008trajectory}:
\begin{equation}
    \pp{B}{O}(s) = \sum_{i=0}^{3} \binom{3}{i} (1-s)^{3-i} s^{i} \p_i.
    \nonumber
\end{equation}
Here, $\p_0$ and $\p_3$ correspond to the current object position and the final target grasp position, respectively. The intermediate control points shape the approach: $\p_1$ is defined along the initial object $X$-axis ($\x_{O}$) to enforce a smooth tangent departure along the initial direction $\boldsymbol{d}_{s}$, while $\p_2$ is obtained by retracting from $\p_3$ along the desired approach direction $\boldsymbol{d}_{a}$, oriented outward from the user's palm to ensure a natural approach. Formally,\\[-2mm]
\[
    \p_1 = \p_0 + \alpha_{s} \boldsymbol{d}_{s} \hspace{5mm} \mbox{ and } \hspace{5mm}
    \p_2 = \p_3 - \alpha_{a} \boldsymbol{d}_{a} 
\]
where $\alpha_s$ and $\alpha_a$ are tunable scaling factors; the resulting position path is shown in Fig.~\ref{fig:bezier}.

The orientation of the object $\R{B}{O}(s)$ is interpolated by \acrfull{slerp} between the initial and target quaternions. To make sure that the orientation stabilizes before reaching the final position, a tunable orientation-lock factor compresses the rotational interpolation within a fraction of the total trajectory duration. This guaranties that the object reaches the desired task-oriented attitude as it approaches the user's hand.\\
Finally, the motion law along the geometric path is defined by a fifth-order polynomial $s(t)$, designed to limit the maximum jerk and ensure a smooth and comfortable motion for the human receiver.\\
Before motion execution, the candidate trajectory is validated. The intended cubic B\'ezier path is discretized into 50 samples and evaluated with a Pinocchio-based inverse-kinematics solver \cite{carpentier2019pinocchio}. A path is accepted only if all samples satisfy the joint-limit margins, a minimum-manipulability bound, and a minimum singular-value bound on the end-effector Jacobian.
If the nominal path fails validation, the planner does not alter the target. Instead, it progressively reduces the scaling factors $\alpha_s,\ \alpha_a \in {1.0,\ 0.75,\ 0.5,\ 0.25,\ 0.0}$ that multiply the offsets of the intermediate control points $\mathbf{p}_1$ and $\mathbf{p}_2$ from the path endpoints, respectively; in the present implementation both factors are reduced jointly along this schedule, though they can be tuned independently. Because the endpoints, the total motion duration, and the minimum-jerk time parameterization are held constant, shrinking the $\alpha$ values flattens the approach curvature without changing the trajectory timing or the normalized velocity profile. The first trajectory that passes validation is executed.
If no scaling factor yields a safe path, the target is rejected, and the system reverts to waiting for a new stable hand pose.
\subsection{Experimental Setup and Task Design}
\label{subsec.ExpSetup_TaskDesign}
The experimental workspace is organized to facilitate repeatable human-robot interactions. The setup features four industrial tools (screwdriver, wrench, screw, and hammer) positioned on the robot's left side, a deposit table, and a predefined handover volume monitored continuously by the Intel RealSense camera to enable real-time hand pose estimation. 
To streamline the interaction, the handover sequence is actively initiated by the user via a clip-on microphone. A speech-to-text model converts the voice command, which is then sent to an LLM (gpt-5-mini) restricted to classifying the request into one of the four available tools. Once chosen, the robot delivers the object. Participants were instructed to place their right hand in the handover volume and receive the tool using a task-specific grasp consistent with its intended use (see Fig.~\ref{fig:grasping_pose}). After the exchange, they must place the object on the deposit table. The procedure was evaluated under both the \textit{Baseline} and \textit{Adaptive} conditions detailed in Section~\ref{subsec.control_traj}.
\subsection{Procedure and Metrics}
\label{subsec:procedure_metrics}
We employed a counterbalanced within-subjects design to mitigate ordering effects, exposing all participants to both the \textit{Baseline} and \textit{Adaptive} handover modalities. Before the experimental sessions, users received a safety briefing and completed two baseline surveys: the Edinburgh Handedness Inventory of 10-items \cite{Oldfield_1971} to confirm right-hand dominance and a 9-item Ethical Acceptability Scale \cite{Peca_2016} (5-point Likert) to assess their \textit{a priori} disposition toward robotics.
The evaluation framework combined objective behavioral and physiological data with subjective questionnaires. During the interaction, participants wore NEON Pupil Labs eye-tracking glasses, from which we extracted the blink rate (blinks/min) to monitor real-time variations in cognitive workload and trust regarding the robot's predictability. Additionally, we measured the \textit{grasping delay}, defined as the elapsed time between the robot's end effector reaching the target handover pose and the user successfully grasping the object. This temporal metric served as an objective proxy for user comfort and interaction fluency, where prolonged delays indicate increased hesitancy or ergonomic discomfort. Subjective perceptions were gathered immediately after each condition. The perceived workload was measured using the NASA-TLX, where participants rated six dimensions on a 20-point scale and made 15 pairwise comparisons to compute a weighted overall score. Finally, interaction safety and reliability were evaluated through a modified 10-item Human-Robot Trust Scale, scored on a 5-point Likert scale (further detailed in Section \ref{subsec:human-robot-trust}, Table \ref{tab:trust_results}).
%

%
\begin{table}[t!]
\centering
\caption{Ethical Acceptability Scale Items: Descriptive Statistics and One-Sample Wilcoxon Test Results}
\label{tab:ethical_acceptability}
\renewcommand{\arraystretch}{1.0}
\begin{tabularx}{\textwidth}{@{} >{\raggedright\arraybackslash}X c c c c c c @{}}
\toprule
\textbf{Ethical Acceptability Statement} & \textbf{Mean} & \textbf{SD} & \textbf{SE} & \textbf{CV} & \textbf{$V$} & \textbf{$p$} \\
\midrule
1. Cobots handing over objects to workers in industrial settings. & 4.53 & 0.64 & 0.17 & 0.14 & 105.00 & < .001* \\
2. Cobots assisting people in domestic environments with everyday objects. & 4.20 & 0.86 & 0.22 & 0.21 & 100.50 & < .001* \\
3. System using cameras and AI models to extract environment info. & 3.40 & 1.30 & 0.34 & 0.38 & 25.00 & .152 \\
4. Handover robots partially replacing human workers in logistics/assembly. & 4.13 & 0.64 & 0.17 & 0.15 & 91.00 & < .001* \\
5. Robots reducing the need for human caregivers in domestic assistance. & 3.33 & 1.11 & 0.29 & 0.33 & 38.00 & .133 \\
6. Human pace of work/life influenced by the robot's reaction times. & 2.73 & 0.88 & 0.23 & 0.32 & 13.50 & .876 \\
7. Delegating physical safety to hardware limits and control algorithms. & 3.00 & 1.31 & 0.34 & 0.44 & 37.50 & .549 \\
8. Programming robot movements to appear ``human'' or ``natural'' for trust. & 3.80 & 1.01 & 0.26 & 0.27 & 67.50 & .010* \\
9. Designing the robotic arm with a friendly/organic appearance. & 3.40 & 1.18 & 0.31 & 0.35 & 39.50 & .104 \\
\bottomrule
\end{tabularx}
\vspace{2pt}
\parbox{\textwidth}{\footnotesize \textit{Note.} $N=15$. One-sample Wilcoxon signed-rank tests evaluated the alternative hypothesis that the median acceptability score was greater than the neutral value of 3. Asymptotic $p$-values are reported. * indicates statistical significance ($p \le .05$).}
\end{table}
\section{Results}
\label{sec.Results}
The analysis combined participants' pre-interaction attitudes toward robotic handover with post-interaction subjective ratings and objective behavioral/physiological measures. The significance threshold was set at $p \le .05$.
\subsection{Pre-study Ethical Acceptability}
Responses to the Ethical Acceptability Scale were analyzed through one-sample Wilcoxon signed-rank tests against the neutral midpoint of 3. As shown in Table~\ref{tab:ethical_acceptability}, participants entered the study with an overall favorable attitude toward robot-mediated handover. Ratings were significantly above the neutral value for the use of cobots in industrial handover scenarios ($M=4.53$, $SD=0.64$, $p<.001$), for domestic object assistance ($M=4.20$, $SD=0.86$, $p<.001$), for the partial replacement of human workers in logistics or assembly tasks ($M=4.13$, $SD=0.64$, $p<.001$), and for the use of deliberately natural-looking motion to increase acceptance and trust ($M=3.80$, $SD=1.01$, $p=.010$). By contrast, the remaining items did not differ significantly from neutrality. Overall, the responses suggest that participants were broadly receptive to the adoption of robotic handover systems, while remaining more cautious on issues involving autonomy, responsibility, and socially persuasive design.

\subsection{Interaction Fluency, Physiological Response and Workload}
\begin{figure}[t]
    \centering
    \includegraphics[width=\linewidth]{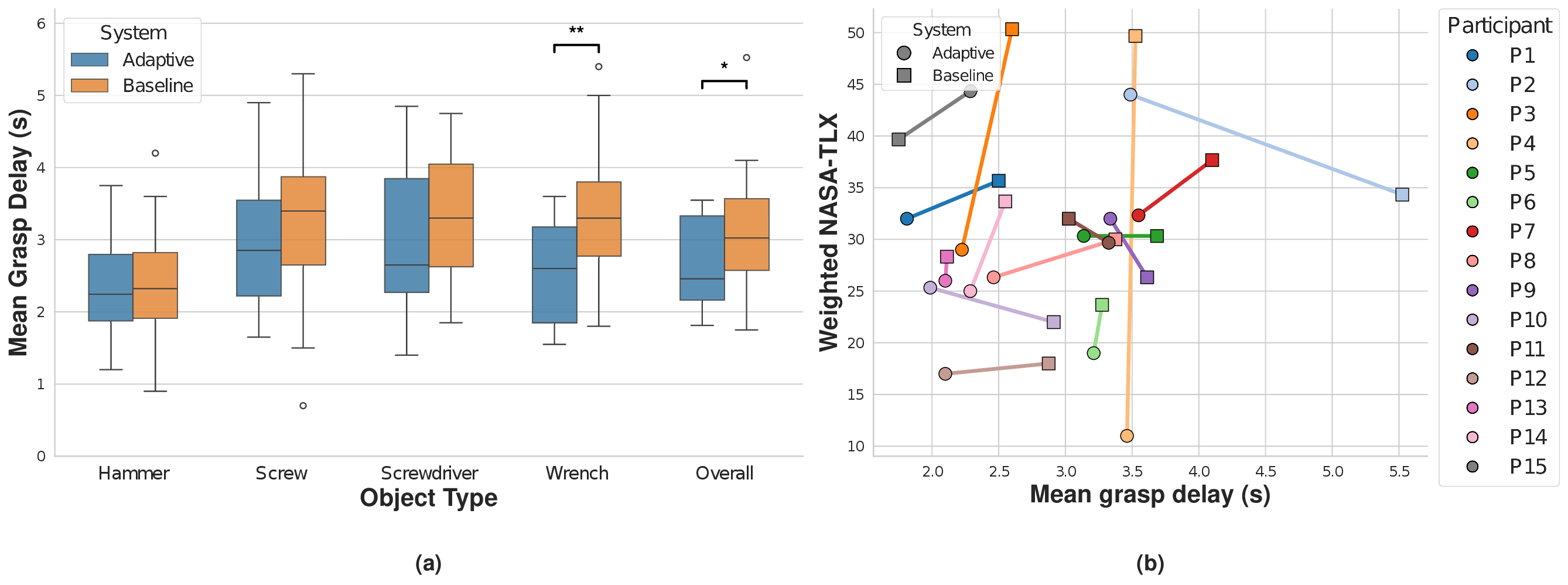}
    \caption{Grasp delay across object types and its relationship with perceived task load.\textit{ (a)} Mean grasp delay per user, divided by object type and overall.\textit{ (b)} Per-participant correlation between mean grasp delay and weighted NASA-TLX score; circles denote the Adaptive system, squares the Baseline.}
    \label{fig:grasp_vs_tlx}
    \vspace{-5mm}
\end{figure}

The clearest effect of the adaptive strategy emerged in the behavioral timing measure. Grasp delay was shorter in the Adaptive condition than in the Baseline condition (Table ~\ref{tab:workload_physiology_results}). Participants grasped the object more quickly with the Adaptive handover ($M=2.72$ s, $SD=0.64$) than with the Baseline handover ($M=3.16$ s, $SD=0.91$), $t(14)=-2.80$, $p=.014$. Since longer delays are interpreted as a proxy for hesitancy or reduced ergonomic comfort, this result indicates that aligning the end-effector orientation to the user's hand improved the fluency of the transfer phase. 
Complementary exploratory analysis by object type is reported in Table~\ref{tab:grasp_delay_by_object}. As portrayed in Fig \ref{fig:grasp_vs_tlx}a, the most pronounced advantage of the Adaptive system was observed for the wrench, for which grasp delay was substantially reduced ($M=2.55$ s, $SD=0.73$) relative to the Baseline condition ($M=3.51$ s, $SD=1.22$), $t(14)=-3.64$, $p=.003$. Screwdriver handovers showed a descriptively similar tendency in favor of the Adaptive condition ($2.97$ s vs. $3.57$ s), although this effect did not reach significance ($p=.108$). No reliable differences emerged for the hammer or screw. Taken together, this pattern demonstrates that orientation adaptation is highly beneficial for tools whose affordances require a strict, directional grip for applying torque. When these tools are handed over at unaligned angles, users are forced to make awkward wrist adjustments that cause measurable hesitation. Aligning the robot's end-effector to the user's natural hand orientation eliminates this ergonomic bottleneck, yielding significantly faster handovers. \\
By contrast, the global measures of workload and physiological response did not show statistically reliable differences between conditions. Weighted NASA-TLX scores were descriptively lower for the Adaptive handover ($M=28.22$, $SD=8.85$) than for the Baseline handover ($M=32.78$, $SD=9.14$), but the paired Wilcoxon test did not reach significance, $W=33.00$, $z=-1.22$, $p=.221$. Similarly, spontaneous blink rate remained comparable across conditions, with $M=18.44$ blinks/min ($SD=7.81$) for the Adaptive system and $M=17.42$ blinks/min ($SD=8.39$) for the Baseline system, $t(14)=1.20$, $p=.249$. Since the reduction in cumulative task load observed for the Adaptive system relative to the Baseline constitutes a first indication
of the superiority of the proposed handover policy, we further
investigated whether the subjective workload co-varied with an
objective measure of interaction fluency. To this end, we investigated how subjective workload visually correlates with objective grasp delay. Fig.~\ref{fig:grasp_vs_tlx} b) suggests that per-user weighted NASA-TLX scores and mean grasp delays co-varied coherently: for most participants, the shift from the Adaptive to the Baseline system pairs longer grasp delays with higher perceived effort. Although isolated within-subject inversions exist, likely due to individual differences in prior robot expertise, the dominant group-level trend ($N=15$) supports the ergonomic and cognitive superiority of the Adaptive handover policy.

\begin{table}[!h]
\centering
\setlength{\tabcolsep}{8pt} 
\renewcommand{\arraystretch}{1.2} 

\caption{Normality and Paired Comparisons for Workload and Physiological Metrics}
\label{tab:workload_physiology_results}

\resizebox{\columnwidth}{!}{%
\begin{tabular}{lcccccc}
\toprule
\multirow{2}{*}{\textbf{Metric}} & \textbf{Adaptive} & \textbf{Baseline} & \textbf{Shapiro-Wilk} & \multirow{2}{*}{\textbf{Test}} & \multirow{2}{*}{\textbf{Statistic}} & \multirow{2}{*}{\textbf{$p$-value}} \\
& \textbf{Mean (SD)} & \textbf{Mean (SD)} & \textbf{$W$ (diff.)} & & & \\
\midrule
Blink Rate & 18.44 (7.81) & 17.42 (8.39) & 0.91 & Paired $t$-test & $t(14)=1.20$ & .249 \\
Weighted TLX & 28.22 (8.85) & 32.78 (9.14) & 0.80 & Wilcoxon & $W=33.0,\ z=-1.22$ & .221 \\
Grasp Delay (s) & 2.72 (0.64) & 3.16 (0.91) & 0.93 & Paired $t$-test & $t(14)=-2.80$ & .014* \\
\bottomrule
\end{tabular}%
}

\vspace{1ex} 
\parbox{\columnwidth}{\footnotesize \textit{Note.} $N=15$. Shapiro-Wilk tests were applied to the paired-difference distributions. Blink rate and grasp delay satisfied the normality assumption sufficiently for paired $t$-tests, whereas weighted TLX did not and was therefore analyzed with a paired Wilcoxon signed-rank test. Grasp delay was computed as each participant's mean over matched object pairs. * indicates statistical significance ($p \le .05$).}
\end{table}


\subsection{Human--Robot Trust}
\label{subsec:human-robot-trust}
Post-study trust ratings were compared item by item using paired Wilcoxon signed-rank tests (Table~\ref{tab:trust_results}). The overall profile was descriptively more favorable for the Adaptive condition.
Two item-level differences reached significance in the analysis. Participants reported greater agreement with the statement that they were not worried because the robot moved as expected when interacting with the Adaptive system ($M = 4.20$, $SD = 0.56$) than with the Baseline system ($M = 3.40$, $SD = 1.18$), $W = 57.00$, $z = 2.23$, $p = .026$. In addition, they reported lower discomfort due to task complexity in the Adaptive condition ($M=1.20$, $SD=0.41$) than in the Baseline condition ($M=1.60$, $SD=0.83$), $W=0.00$, $z=-2.12$, $p=.034$. These findings indicate that the adaptive orientation strategy made the robot's behavior easier to interpret and the handover easier to manage from the user's perspective.
The remaining trust items did not reach the significance threshold, although several of them showed descriptively more favorable scores for the Adaptive condition.
Overall, the results identify an advantage for the Adaptive handover. Coherent adaptation of the end-effector orientation to the receiver's hand did not significantly alter blink rate or global workload, but it reduced grasp delay and selectively improved trust dimensions linked to motion expectancy and interaction simplicity. In practical terms, this suggests that the adaptive policy improves how users commit to the grasp once the robot reaches the handover pose.

\begin{table}[!t]
\centering
\setlength{\tabcolsep}{8pt} 
\renewcommand{\arraystretch}{1.2} 

\caption{Exploratory Paired Comparisons for Grasp Delay by Object}
\label{tab:grasp_delay_by_object}
\resizebox{\columnwidth}{!}{%
\begin{tabular}{lccccc}
\toprule
\textbf{Object} & \textbf{$N$} & \textbf{Adaptive Mean (SD)} & \textbf{Baseline Mean (SD)} & \textbf{Paired $t$-test} & \textbf{$p$-value} \\
\midrule
Hammer & 15 & 2.35 (0.74) & 2.33 (0.93) & $t(14)=0.08$ & .937 \\
Screw & 15 & 3.01 (0.92) & 3.22 (1.14) & $t(14)=-0.49$ & .628 \\
Screwdriver & 15 & 2.97 (1.01) & 3.57 (1.49) & $t(14)=-1.72$ & .108 \\
Wrench & 15 & 2.55 (0.73) & 3.51 (1.22) & $t(14)=-3.64$ & .003* \\
\bottomrule
\end{tabular}%
}

\vspace{1ex} 
\parbox{\columnwidth}{\footnotesize \textit{Note.} * indicates statistical significance ($p \le .05$).}
\end{table}

\begin{table*}[h!]
\centering
\caption{Descriptive Statistics and Wilcoxon Signed-Rank Test Results for the Human--Robot Trust Scale}
\label{tab:trust_results}
\footnotesize
\setlength{\tabcolsep}{3.8pt}
\renewcommand{\arraystretch}{1.08}
\begin{tabularx}{\textwidth}{@{}>{\raggedright\arraybackslash}X c c c c c c c@{}}
\toprule
\multirow{2}{*}{\textbf{Trust Scale Item}} & \multicolumn{2}{c}{\textbf{Adaptive}} & \multicolumn{2}{c}{\textbf{Baseline}} & \multicolumn{3}{c}{\textbf{Wilcoxon Test}} \\
\cmidrule(lr){2-3} \cmidrule(lr){4-5} \cmidrule(lr){6-8}
& \textbf{Mean} & \textbf{SD} & \textbf{Mean} & \textbf{SD} & \textbf{$W$} & \textbf{$z$} & \textbf{$p$} \\
\midrule
The way the robot moved made me uncomfortable & 1.47 & 0.74 & 1.87 & 1.13 & 2.50 & -1.73 & .084 \\
I wasn't worried because the robot moved as I expected & 4.20 & 0.56 & 3.40 & 1.18 & 57.00 & 2.23 & .026* \\
The robot's speed made me uncomfortable & 1.27 & 0.59 & 1.53 & 1.06 & 0.00 & -1.63 & .102 \\
Speed of gripper picking/releasing made me uneasy & 1.60 & 0.99 & 1.60 & 1.12 & 5.00 & 0.00 & 1.000 \\
I felt I could rely on the robot to perform its tasks & 4.07 & 0.59 & 3.73 & 0.88 & 24.00 & 1.89 & .059 \\
I knew the gripper wouldn't drop the objects & 4.07 & 0.70 & 3.73 & 1.10 & 18.00 & 1.67 & .096 \\
I felt safe interacting with the robot & 4.40 & 0.63 & 3.87 & 0.99 & 19.00 & 1.81 & .071 \\
I felt confident cooperating with the robot & 4.33 & 0.62 & 3.93 & 1.03 & 18.50 & 1.73 & .084 \\
I felt uncomfortable cooperating due to task complexity & 1.20 & 0.41 & 1.60 & 0.83 & 0.00 & -2.12 & .034* \\
If the task was more complicated, I'd be more worried & 2.40 & 1.12 & 2.87 & 1.06 & 10.50 & -1.47 & .142 \\
\bottomrule
\end{tabularx}

\vspace{2pt}
\parbox{\textwidth}{\footnotesize \textit{Note.} $N=15$. Two-sided paired Wilcoxon signed-rank tests compare Adaptive and Baseline scores for each item. $W$ denotes the positive-rank sum. Asymptotic $p$-values are reported. No multiple-comparison correction was applied. * indicates statistical significance ($p \le .05$). Responses were recorded on a 5-point Likert scale.}
\end{table*}

\section{Conclusion}
\label{sec:conclusion}
This study presented an adaptive, voice-driven handover system for mechanical tools built on a Franka Panda robot setup. By integrating real-time 3D hand pose detection via MediaPipe with an LLM to classify user voice requests, the framework modulates the end-effector orientation to present the tool in an ergonomically optimal pose for the receiver.
Experimental results demonstrate that this adaptive strategy improves interaction fluency for specific orientation-sensitive tools compared to a static, task-agnostic baseline. Behavioral metrics revealed a significant reduction in grasp delay, with the most pronounced advantage observed for tools like the wrench that require a strict, directional grip for applying torque. By presenting tools in an optimal, handle-first orientation, the framework eliminates ergonomic bottlenecks and the resulting hesitancy. Participants reported that the adaptive system's movements were more aligned with their expectations and reduced discomfort related to task complexity, selectively improving dimensions of human-robot trust.
This work highlights the potential of combining intent recognition with reactive kinematic control and adaptation in collaborative manufacturing cells. Implementing adaptive handover policies may effectively resolve ergonomic bottlenecks and reduce grasping delays, paving the way for safer, more efficient, and highly fluent shared autonomy in industrial environments.
Future work will focus on expanding the user study with more candidates, employing a multi-camera setup to achieve a more robust and accurate view of the handover area. Additionally, we plan to integrate hand-tracking gloves to directly assess ergonomic comfort and physical fatigue during the exchange phase. In order to ehnahce the framework, we aim to incorporate an artificial intelligence model capable of dynamically estimating the optimal grasping frame based on the specific object requested, further extending the system's adaptability. Finally, future investigations will explore the robot's capacity to predict human intent in inherently ambiguous, complex collaborative scenarios, during shared assembly tasks where the robot must proactively select its role and synchronize force and timing with the operator.

\newpage
\begin{credits}
\subsubsection{\ackname}
This work was partially supported by the Italian National Recovery and Resilience Plan (PNRR), Mission 4 “Education and Research”, Component C2, Investment 1.1 “PRIN – Projects of Relevant National Interest”, Project I-SHARM: Intelligent SHared Autonomy for Robotic Manipulation Systems, Project ID 2022NTZRFM, CUP E53C24002600006 and by the University of Modena and Reggio Emilia under the FAR (Fondo di Ateneo per la Ricerca – Linea Fomo) project titled \textit{ROBIN3: a ROBotic INTelligent, INTuitive, and INTeractive platform for NAO-Mediated Autistic Healthcare}.

\subsubsection{\discintname}
The authors have no competing interests to declare.
\end{credits}

\bibliographystyle{IEEEtran}
\bibliography{references}

\end{document}